\documentclass[2019, 0]{PHMSociety}
\usepackage{amsmath,bbm,graphicx}
\usepackage{multirow}
\usepackage{color}
\usepackage{caption}
\usepackage{subcaption}
\usepackage{wrapfig}
\usepackage{nowidow}
\usepackage{url}

\usepackage{flushend}
\usepackage{apacite}

\newcommand{\p}[1]{\text{P}\left( #1 \right)}

\newcommand\norm[1]{\left\lVert#1\right\rVert}
\newcommand{\paren}[1]{\left( #1 \right)}
\newcommand{\keyword}[1]{\textit{#1}}

\begin{document}

% Paper Title
\title{Multi-label Prediction in Time Series Data \\using Deep Neural Networks}
%A Neural Networks Approach to Multi-label Time Series Prediction \\With Severe Class Imbalance
\author{%			
    Wenyu Zhang\authorNumber{1}, Devesh K. Jha\authorNumber{2}, Emil Laftchiev\authorNumber{2}, and Daniel Nikovski\authorNumber{2}
}

% Author Affiliations
\address{% This is a tabular environment so each affiliation needs to be separated by "\\" or "\tabularnewline"
	\affiliation{1}{Department of Statistics, Cornell University, Ithaca, NY, USA}{ %add emails
		{\email{wz258@cornell.edu}}
		} % emails input
	\tabularnewline % skip one row for next affiliation	
	\affiliation{{2}}{Mitsubishi Electric Research Labs (MERL), Cambridge, MA, USA}{ % add emails
		{\email{\{jha, laftchiev, nikovski\}@merl.com}}
		} % emails input
}

% Create the title
\maketitle

% PHM Society Distribution License Information, provide first author's name "FirstName LastName"
%\phmLicenseFootnote{Wenyu Zhang}

\begin{abstract}
This paper addresses a multi-label predictive fault classification problem for multidimensional time-series data. While fault (event) detection problems have been thoroughly studied in literature, most of the state-of-the-art techniques can't reliably predict faults (events) over a desired future horizon. In the most general setting of these types of problems, one or more samples of data across multiple time series can be assigned several concurrent fault labels from a finite, known set and the task is to predict the possibility of fault occurrence over a desired time horizon. This type of problem is usually accompanied by strong class imbalances where some classes are represented by only a few samples. Importantly, in many applications of the problem such as fault prediction and predictive maintenance, it is exactly these rare classes that are of most interest. To address the problem, this paper proposes a general approach that utilizes a multi-label recurrent neural network with a new cost function that accentuates learning in the imbalanced classes. The proposed algorithm is tested on two public benchmark datasets: an industrial plant dataset from the PHM Society Data Challenge, and a human activity recognition dataset. The results are compared with state-of-the-art techniques for time-series classification and evaluation is performed using the F1-score, precision and recall.\\
\keyword{keywords: Time-series analysis, Fault Detection, Fault Prediction.}
\end{abstract}

\section{Introduction}
\label{sec:intro}

Time series analysis for rare events such as faults is generally a known and difficult problem~\cite{yamanishi2002unifying}. The problem is particularly difficult in the multi-dimensional, or multi-variate setting, where the events may be described by simultaneous occurrences on multiple time series. A key confounding factor is the number of class labels, or the number of events that must be discovered. In the worst case, the events are described by labels from a known set, but may occur simultaneously. Naturally, this approach to labeling leads to a number of classes that grows combinatorially with the number of individual labels in the original set. The problem in this setting is one of time series analysis on infrequent events with multi-label classes that are characterized by severe class imbalance. 

This is a critical problem in the field of prognostics and health management where faults for an equipment occur rarely but often have serious impacts such as service disruption, safety concerns for users and associated costs of repair or replacement. In these fields, the detection of rare events (fault detection, anomaly detection) in a timely fashion is critical to minimizing the detrimental impacts of the events. An example of this type of application was demonstrated in Holst et. al. \cite{holst2012} where a fault detection tool was used to on a fleet of trains leading to reduced long-term maintenance costs by $5-10\%$.

While detecting faults is important~\cite{chandola2009anomaly, 8795698} and many models have been used including deep learning~\cite{jha2016temporal, sarkar2016multimodal}, today the field of anomaly detection is moving towards predicting when a fault will occur. This capability is commonly known as prognostics, or the ability to predict impending faults. Prognostics is a critical technology in the drive to reduce cost of operation. The key difference is that while anomaly detection leads to corrective maintenance, fault prediction leads to predictive maintenance which can be performed before failures occur or before they develop into major defects. In practice this means that the condition of the equipment is monitored, and the time to failure is continuously estimated. Such an approach is able to reduce the number of repairs, allow for optimal scheduling of maintenance workers and improve the safety of human operators. %\cite{maintenance2018}.

This paper focuses on the problem of fault prediction in the multidimensional time series case with multi-label classes some of which a have few examples. We propose a recurrent neural network-based (RNN) approach that in addition to the observed time series incorporates contextual information for past, present and future time intervals. This contextual information describes the operation of the plant/machine and could contain information about the target set-point, type of control being used for the plant, expected output, speed of plant/machine operation, etc. Using this approach, we are able to predict data labels for future time segments and to localize the faults that could occur in future with better accuracy than several other state-of-the-art algorithms for time-series analysis. 
\begin{figure}
    \centering
    \includegraphics[width=.8\linewidth]{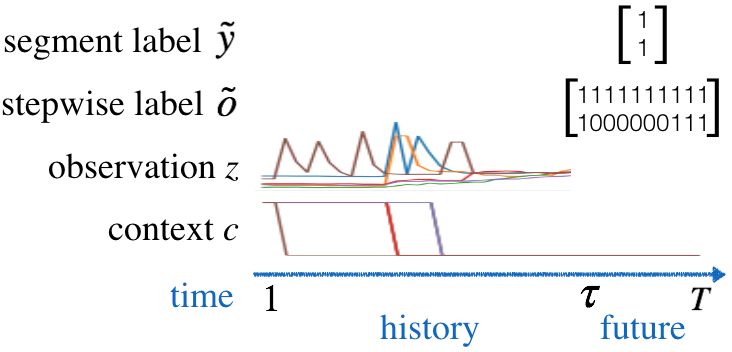}
    \caption{Notation used in this paper. Example given in the setting of industrial plant fault prediction with two fault types: binary labels denote the presence of fault, observations correspond to sensor measurements, context sequence are control system setpoints.}
    \label{fig:variables}
\end{figure}

Figure \ref{fig:variables} illustrates the problem setting and the paper notation. Here $\tau$ is the length of observed data, and $T-\tau$ is the prediction horizon. For a sample $i$ of some monitored system, $z^{(i)}_{1:\tau} = \left[ z^{(i)}_1, \dots, z^{(i)}_\tau \right]$ of length-$\tau$ is a historical sequence of observations where $z^{(i)}_t \in \mathbbm{R}^{d_z}$. During the historical period, there are no labels that indicate the state of faults. 
For the future time segment, fault labels are predicted at two levels of granularity through segment labels and stepwise labels. Given $L$ possible fault labels, $\tilde{y}^{(i)} \in \mathbbm{R}^L$ denotes the segment label where each element $\tilde{y}^{(i)}[\ell] \in \{0, 1\}$ indicates the absence or presence of a label $\ell$ in the next $T-\tau$ steps. For example, if $L=2$, the set of possible segment labels is $\left\lbrace [0,0], [1,0], [0,1], [1,1] \right\rbrace$. The stepwise label is $\tilde{o}^{(i)}_{\tau+1:T} = \left[ \tilde{o}^{(i)}_{\tau+1}, \dots, \tilde{o}^{(i)}_T \right]$ where each binary vector $\tilde{o}^{(i)}_t \in \mathbbm{R}^L$ indicates the absence or presence of the $L$ labels at step $t \in \{\tau+1, \dots, T\}$. Here $\tilde{o}^{(i)}_{\tau+1:T}$ provides label localization on top of $\tilde{y}^{(i)}$. Mathematically, $\tilde{y}^{(i)}[\ell] = \mathbbm{1}\{ \sum_{t=\tau+1}^T \tilde{o}^{(i)}_t[\ell] > 0 \}$.
There is no restriction on the number of one's in the vectors $\tilde{y}^{(i)}$ and $\tilde{o}^{(i)}_t$, as is consistent with the definition of a multi-label classification problem. In the fault prediction context, this means that multiple types of faults are allowed to occur simultaneously. The case where multiple components of a system are monitored for failure is also included. 

Contextual input $c^{(i)}_{1:T} = \left[ c^{(i)}_1, \dots,  c^{(i)}_T \right]$ where $c^{(i)}_t \in \mathbbm{R}^{d_c}$ can include any independent variable of the system that is fully known at time $\tau$ (ex. pre-specified setpoint control sequences in industrial plants - Figure \ref{fig:variables}.) The presence of a label $\ell$ is a function of both input $z$ and $c$ without further assumptions on the structure within the inputs. 

In this paper, we would like to estimate $\p{\tilde{y}[\ell] = 1 | z_{1:\tau}, c_{1:T}}$, the probability of  given label in a future segment. To estimate this probability, the usual approach is to first define two functions $F$ and $V^\ell$ on the input variables $z_{1:\tau}, c_{1:T}$. The function $V^\ell$ transforms the inputs from the observed space into an embedded space, while the function $F$ maps from the embedded space to the interval $[0,1]$. A simple map is chosen so that there is a clear and interpretable relationship between embedding and the output. Second, the optimal parameters of these functions are learned such that $F\left( V^\ell \left( z_{1:\tau}, c_{1:T} \right) \right)$ approximates $\p{\tilde{y}[\ell] = 1 | z_{1:\tau}, c_{1:T}}$ as closely as possible. 

% The function $V^\ell$ usually embeds the observed data in a space that is lower dimensional than the input space. This is because the original data is often sparse and $V^\ell$ in this case becomes a mappping from the high dimensional space to a lower-dimensional non-linear manifold. In this paper, the function $V^\ell$ consists a RNN plus further arithmetic manipulations on the RNN hidden units and maps the input variables to $\mathbbm{R}$. The low dimensional embedding space is further helpful because it allows straightforward visualisation of the embedded representations.
%suitable loss function for multidimensional time-series data under severe class imbalance
This paper represents a new neural network architecture which we call Multi-label Predictive Network (MPN) using RNNs with a loss function that is designed considering the class imbalance. The architecture is fashioned after the sequence-to-sequence (Seq2Seq) model which has been successful in neural machine translation \cite{sutskever2014}. This paper demonstrates that the model can be adapted for a different objective of multi-label prediction for temporal data. %The main insight that we present here is that for prediction of multiple possible faults, instead of predicting the future time series (which could be error prone for the high dimensional data), we predict the future possible faults one time instant at a time. The predictions are then fed-back to the network along with the available contextual information to make predictions further ahead in time.

We follow the insight that while prediction of multi-dimensional time-series data is a complex task due to the high-dimensional temporal data and the complex correlations which might be present between the different components of the data. It is, however, easier to learn the dynamics of the faults occurring in the system using a low-dimensional representation of the time-series data in an abstract space. 

Neural networks used for sensor data compression are shown to be capable of capturing intramodal and intermodal correlations, and the compressed representations are effective for prediction and missing data imputation tasks \cite{liu2017}. In prognostics and health management, neural networks can learn useful features to estimate remaining useful life and degradation \cite{li2018}.

For accurate prediction of faults over a certain prediction horizon, the faults predicted by the network at every instant of time should be taken into account as the faults may be correlated and thus prediction of a fault at a time-instant '$k$' might affect of the probability of occurrence of the fault at a future instant '$k+1$'. Additionally, this information could also be useful in the multi-fault setting as some faults could always occur together or some faults never occur together and thus, this can simplify learning of the fault dynamics. As a result of this motivation, the proposed MPN predicts faults one step at a time, and then the predicted fault is fed back to the network for making predictions on the faults occurring in future over a finite window of time.

% In this paper MPN models the temporal dynamics in the multidimensional time series and is suitable for applications such as fault prediction in the case of multilabel classes with rare events. 
The loss function used to train MPN is comprised of sub-objectives that predict the true segment labels and the stepwise labels, $\tilde{y}$ and $\tilde{o}$. Both label types are multi-label, which means that the number of possible labels scales with the number of discrete labels in the label set. Samples can thus be labeled with any number of labels ranging from none (healthy) to all possible labels in the label set. For each label class, the loss is weighted to account for the imbalance in the training data set. Classes that are rare are weighted more heavily. The loss function also includes regularization components to (possibly) prevent overfitting in the network. 
The proposed ideas for fault prediction using MPN are verified using an industrial plant dataset from the 2015 Prognostics and Health Management Society (PHM Society) Data Challenge which consists a total of 33 plants. The network is also tested on a second dataset which is a human activity recognition (HAR) dataset from the 2011 Opportunity Activity Recognition Challenge, and we aim to predict low-level motions based on sensor readings and high-level activities. Both datasets are publicly available. Comparisons are conducted with several state-of-the-art techniques for time series classification, and MPN is shown to have better performance in general in terms of precision, recall and F1-score. Compared to the best performers in the two data challenges which use hand-engineered features, MPN attains comparable or better performance without the need for extensive feature extraction.

\section{Related Work}
\label{sec:related}

An important quantity for predictive maintenance is the remaining useful life (RUL). RUL of an equipment is the length from the current time to the end of its useful life, which is usually a unit of time, but can also be other measurements such as revolutions for rotating machines, number of operations for switchgear and load cycles for structural components \cite{welte2014}. If a model of the performance degradation path can be built, RUL can also be found by determining the time at which degradation exceeds the failure threshold \cite{hu2015}. The rate of degradation is usually not known in real world applications where physical simulation technology or accelerated life testing is not possible. In general, physical and stochastic models may not sufficiently capture the dynamics in the data from a complex system. RUL estimation does not reflect the duration of a fault, while the proposed MPN also attempts to localize the start and end times of faults in a target window.

A conventional data-driven approach for label prediction is to directly perform classification on the historical observations with standard shallow classifiers \cite{salfner2010}. Ensemble-based fault detection methods frequently use hand-engineered features \cite{xiao2016, xie2016} that are hard to identify and require domain knowledge. The label power set strategy represents multi-label combinations as distinct classes \cite{tsoumakas2011} however this approach is limited by unobserved classes and exponential space and computational complexity. The simplifying assumption of independent labels allows for training a separate binary classifier for each label. Classifier chains \cite{read09} can be used to preserve information on label dependence, but chain order can significantly affect performance. A long short term memory (LSTM) RNN on raw time series with replicated targets achieves state-of-the-art performance in medical diagnoses \cite{lipton2015}. In this paper, LSTM models are used in a predictive network to take advantage of their ability to model long-term dependencies. 

Alternative distance-based approaches include Shapelet Forests (S-F) \cite{patri2014} which extracts subsequences that are the most discriminative for each class and calculates their distances to the time series for classification. K-Nearest Neighbors (KNN) approaches use a majority vote of the neighbors to classify a sample. Dynamic Time Warping (DTW) distance can quantify the qualitative dissimilarity of sequences and is therefore often used for human activity recognition \cite{seto2015}. Siamese LSTM networks learn a similarity metric from time series by first embedding sample pairs in a new space and then comparing the embedded representations using cosine similarity \cite{pei2016}. All KNN methods depend on the diversity and quality of the training samples which is a significant issue in the setting of severe class imbalance.

When the data is fully labeled, Markov models combined with an iterative strategy can be used to do multi-step forward prediction \cite{read2017}. A common approach is time series prediction followed by classification \cite{molaei2015}. However such labeling is generally not available in real applications.

The Seq2Seq model uses two LSTM networks to generate output sequences for neural machine translation \cite{sutskever2014}. Extensions of RNNs have benefited from including spatial and temporal information \cite{liu2016}. RNNs have also seen success in RUL estimation, with the benefit that a neural network setup can be configured to include contextual information such as workloads, operating conditions and deterioration modes \cite{zheng2017, heimes2008} In this paper, the Seq2Seq model is modified to incorporate temporal contextual input in the forecasting window and to output multi-label predictions. Moreover, the approach here avoids sequential error sources by directly predicting time series labels.

\section{Proposed Approach}
\label{sec:model}

The objective of this work is to predict fault labels in a multidimensional time series on a window of time in the future. This prediction is performed using historical observations of the time series and contextual (application specific) input. Given $L$ labels, the proposed approach has two steps. First, using MPN, we learn an $L$-dimensional predictive representation of the data in the future time segment. Then second, we use a linear classifier in each dimension of the predictive representation to determine the presence or absence of a given label.  
Importantly, each dimension $\ell$ of the representation corresponds to a label $\ell$ in the total label set and all dimensions are learned jointly. The linear classification is thus performed per dimension and independently of the other dimensions. This independence in the prediction means that the number of dimensions is equal to the number of faults and not the size of the power set which is all possible combinations of faults.

To reduce the complexity of the predictive problem, the MPN predicts each fault separately. This is motivated by the realization that the network learns the coupled, correlated dynamics between different faults during training. This is because the network is trained to predict faults one step at a time and these predictions are then used to predict faults in the future. In such a scenario, the network learns the dependence of faults occurring in time-- more specifically, it can possibly learn which faults act as precursors to other faults. As a result of this simplification, the dimension of the predicted vector is reduced to the number of faults occurring in the system instead of all the possible combinations of faults in the system. This simplification also allows to alleviate the sample complexity of the neural network as it has to learn fewer classes.
%In effect, this approach uses the fact that neural networks are universal function approximators.

The predictive representation is modeled using a neural network because neural networks have the advantage of being able to model irregular data shapes and complex system dynamics from data. The network in this paper consists of two connected LSTM networks: the encoder and decoder networks. Each network is turned to capture the time series dynamics in either the historical data (network 1) or the future segment data (network 2). The following equations describe the update step of an LSTM network. Here at each step, given input $x_t$ and network states at the previous time step, the current hidden state $h_t$ is obtained as \cite{lstm2015}: 
%\begin{centermath}
\begin{align}
f_t &= \sigma \left( W_f \cdot \left[h_{t-1}, x_t\right] + b_f\right) \nonumber\\
i_t &= \sigma \left( W_i \cdot \left[h_{t-1}, x_t\right] + b_i \right) \nonumber\\
\tilde{\xi}_t &= \tanh\left( W_\xi \cdot \left[h_{t-1}, x_t\right] + b_\xi \right) \nonumber\\
\xi_t &= f_t \odot \xi_{t-1} + i_t \odot \tilde{\xi}_t \nonumber\\
q_t &= \sigma \left( W_q \cdot \left[h_{t-1}, x_t\right] + b_q\right) \nonumber\\
h_t &= q_t \odot \tanh\left(\xi_t\right)
\end{align}
%\end{centermath}
\noindent where $\sigma$ is the sigmoid function, $\tanh$ is the hyperbolic tangent and $\odot$ is the Hadamard product. The cell state is $\xi_t$. The forget gate is denoted as, $f_t$, the input gate is denoted as $i_t$, and the output gate is denoted as $q_t$. The hidden states undergo additional computations to arrive at the final output of the network. These computations are application specific. The weight matrix $W$ and bias vector $b$ corresponding to each gate are learned to minimize the loss function of the network. For $h_t \in \mathbbm{R}^L$ and $x_t \in \mathbbm{R}^m$, the total number of trainable parameters for an LSTM network is $4\left(L^2 + Lm + L\right)$. The MPN encoder is an one-layer LSTM network with $L$ units, $x_t$ is a concatenation of $z_t$ and $c_t$, and $m = d_z + d_c + L$. The MPN decoder is another one-layer LSTM netwok with $L$ units, $x_t$ is $c_t$ and $m = d_c + L$.

\subsection{Predictive Data Representations}
\label{sec:representations}
\begin{figure*}
    \centering
    \includegraphics[width=.75\textwidth,height=9cm]{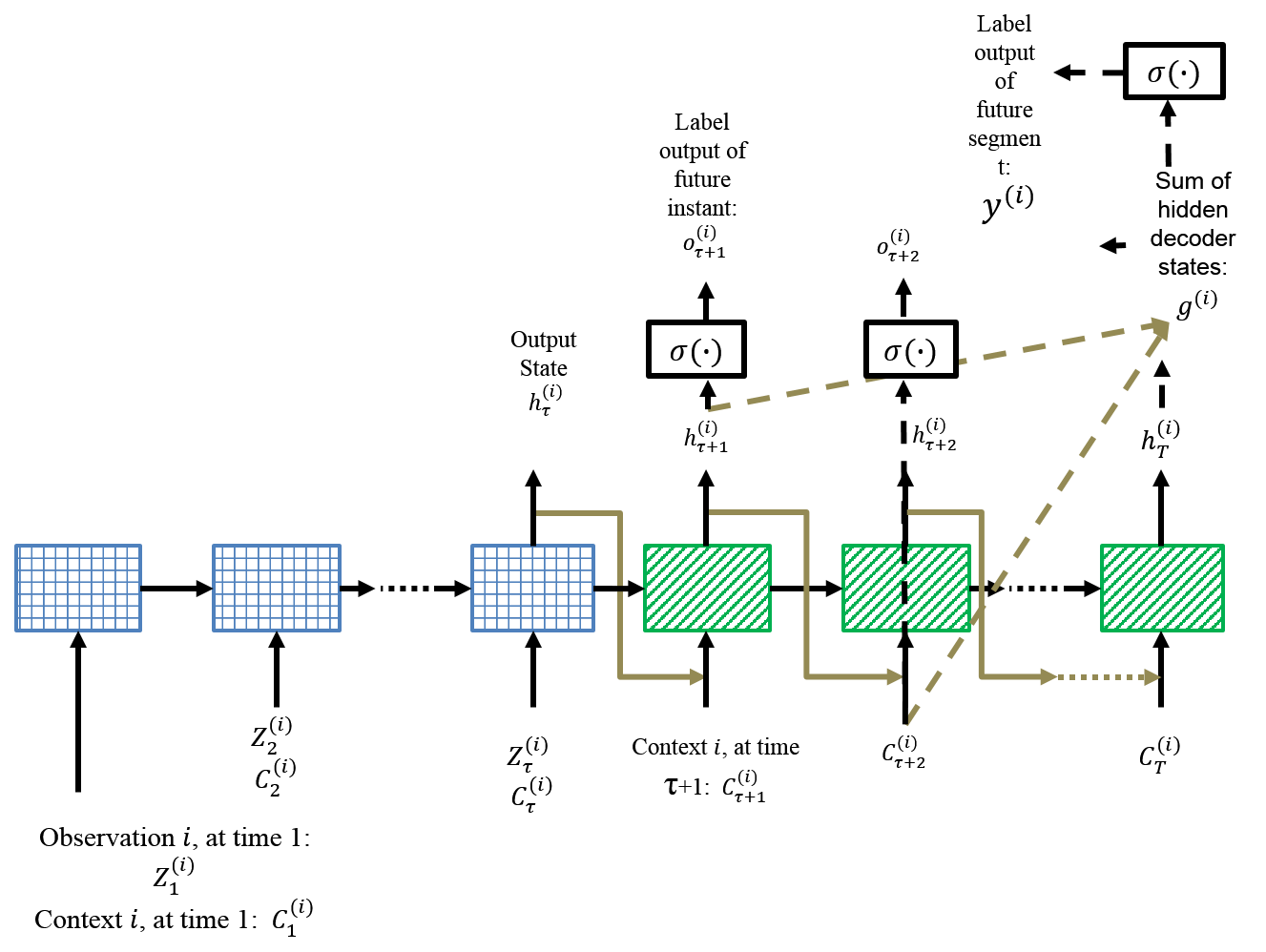}
    \caption{Multi-label Predictive Network (MPN) architecture to embed sample $i$ as $g^{(i)}$. Encoding network (blue square pattern) encodes historical observations $z^{(i)}_{1:\tau}$ and historical context $c^{(i)}_{1:\tau}$, decoding network (green stripped pattern) predicts for labels with additional contextual input $c^{(i)}_{\tau+1:T}$. Hidden units are denoted $h^{(i)}_t$. MPN outputs multi-label predictions $y^{(i)}$ for the segment labels and $o^{(i)}_{\tau+1:T}$ for the stepwise labels.}    
    \label{fig:nn_pred}
\end{figure*}

As previously noted, here we are interested in estimating the conditional probability $\p{\tilde{y}^{(i)}[\ell] = 1 \left| z^{(i)}_{1:\tau}, c^{(i)}_{1:T} \right.}$. This estimate is preformed using MPN which is depicted in Figure \ref{fig:nn_pred}. In this figure, the blue blocks (square pattern) represent the encoder LSTM network. This network encodes the historical data into the embedded space. The green (stripped pattern) blocks represent the decoder LSTM network which decodes the embedded representation into predictions of the labels on a future segment. The flow of information both within the networks and between the two networks is represented by the arrows. Interpreting the figure from left to right, the encoder network synthesizes historical observations $z^{(i)}_{1:\tau}$ and contextual variables $c^{(i)}_{1:\tau}$ to output $h^{(i)}_\tau \in \mathbbm{R}^L$, an estimated representation of labels present at $\tau$. This representation is input to the decoder network together with the final hidden state of the encoder network. Given this synthesized historical information from the encoder network and the future context $c^{(i)}_{\tau+1:T}$, the decoder network directly learns $h^{(i)}_t$ to predict labels for each future time step. 

In the final step, the estimate of $y^{(i)}$ is produced by adding bias to the sum of $h^{(i)}_t$ terms (hidden states) and taking the sigmoid transformation on the total sum.
%\begin{small}
%\begin{centermath}
\begin{align}
    g^{(i)} &= \sum\limits_{t=\tau+1}^T h^{(i)}_t + b_g \\
    y^{(i)} &= \sigma\left( g^{(i)} \right)
\end{align}
%\end{centermath}
%\end{small}
The final output, $y^{(i)}[\ell]$, is the estimate of the desired conditional probability $\p{\tilde{y}^{(i)}[\ell] = 1 \left| z^{(i)}_{1:\tau}, c^{(i)}_{1:T} \right.}$. 
That is, using notations in Section \ref{sec:intro}, the function $V^\ell$ is the first portion of the proposed MPN that maps to $\mathbbm{R}$ to produce the embedding $g$, and the function $F$ is the sigmoid function.

\subsubsection{Designing the Loss Function}
The network described in the last section are trained using a loss function considering two main objectives. Specifically, the loss function has two components: the component which penalizes errors in labeling the whole future time segment, and a component which penalizes errors in labeling each step in the future segment. 

The first component is the component which focuses on learning the labels on the whole future segment. A variable in this component of the objective function is $p_\ell$. $p_\ell$ is the probability of label occurrence in the training set, and it is used as a weight in the cross-entropy loss, $w_\ell = -\log\left( p_\ell \right)$ \cite{jain2016}. Using this weighting is important because it accentuates rare classes and forces the network to learn classes which have very few examples. This weighting is known to increase the false positive rate in the rare classes. However, in this application this increase is acceptable as compared to the relative importance of true positive and the importance of preserving label diversity in the predictions. 

We note that there are many different weight schemes \cite{jain2016}. The choice in this paper is motivated by the empirical experiments which show that this choice optimizes the performance of MPN. Edge cases where $p_\ell = 0$ are assigned a weight of 1 to indicate no preference in predicting label $\ell$ as either present or absent in the test set. The label prediction loss is then expressed as follows with true labels denoted by \textasciitilde (tilde),

%\begin{centermath}
\begin{align}
    & L^{(i)}_Y = - \sum\limits_{\ell=1}^L \left[ w_\ell \tilde{y}^{(i)}[\ell] \log\left( y^{(i)}[\ell] \right) + \right. \nonumber\\
    & \hspace{3cm} \left. \left(1-\tilde{y}^{(i)}[\ell]\right) \log\left( 1-y^{(i)}[\ell] \right) \right]
\end{align}
%\end{centermath}

% The Siamese structure has also been used to address class imbalance by training with sample pairs to learn the similarities and differences between samples \cite{pei2016}. Careful selection of sample pairs is required to ensure that all classes are adequately represented and that the samples are diverse enough. This is particularly difficult to achieve for multi-label tasks due to the many combinations of labels a sample can have. Including a fixed number of samples from each label combination may be infeasible since some combinations may be rarely or never observed in the training set. A Siamese version of MPN is included in the implementations in Section \ref{sec:applications}, it performed similarly as versions without the Siamese structure.
The second component which penalizes errors in labeling each step in the future segment. This component of the objective function uses $o^{(i)}_t = \sigma\left( h^{(i)}_t \right)\sigma\left( g^{(i)} \right) \in (0,1) $, the step wise label vector which annotates the occurrence of labels in each of future segment time steps. At each time instant in the future segment, $o^{(i)}_t[\ell]$ approximates the probability $\p{\tilde{o}^{(i)}_t[\ell] = 1 \left| z^{(i)}_{1:\tau}, c^{(i)}_{1:T} \right.}$ that the label is present or absent at $t$ for sample $i$. 

When $\tilde{o}^{(i)}_t[\ell] = 1$, both $\sigma\left( h^{(i)}_t \right)$ and  $\sigma\left( g^{(i)} \right)$ are $\approx 1$. When a given label is present at $t$ and $h^{(i)}_t[\ell]$ is negative, this function penalizes the result. Similarly, negative value of $g^{(i)}[\ell]$ are penalized when label $\ell$ is present in the forecasting window. Therefore $o^{(i)}_t[\ell]$ tends to be positive for a present label and close to zero for an absent label. $o^{(i)}_t[\ell]$ is used in a squared loss function that ensures training stability.

% \begin{centermath}
% \begin{array}{ll}
%      L^{(i)}_O = \frac{1}{T - \tau} \sum\limits_{t=\tau+1}^T \norm{ \tilde{o}^{(i)}_t - o^{(i)}_t}^2_2
% \end{array}
% \end{centermath}
%\begin{centermath}
\begin{align}
    & L^{(i)}_O = \frac{1}{L(T - \tau)} \sum\limits_{t=\tau+1}^T \left[ \tilde{o}^{(i)}_t \cdot \left(1-o^{(i)}_t\right)^2 + \right. \nonumber\\
    & \hspace{4cm} \left. \left(1-\tilde{o}^{(i)}_t\right) \cdot \left(o^{(i)}_t\right)^2 \right]
\end{align}
%\end{centermath}
Note that the summand is similar to $\norm{ \tilde{o}^{(i)}_t - o^{(i)}_t}^2_2$. The true step wise label $\tilde{o}^{(i)}_t[\ell] \in \{0,1\}$. When $\tilde{o}^{(i)}_t[\ell] = 1$, the loss incurred is $\left(1-o^{(i)}_t[\ell]\right)^2$, and when $\tilde{o}^{(i)}_t[\ell] = 0$, the loss incurred is $\left(0-o^{(i)}_t[\ell]\right)^2$. We write $L_O^{(i)}$ as such so that it has similar form as $L_Y^{(i)}$.

The objective function further has an $\ell_2$ regularization on the weights of the network, collectively denoted as $W$, to prevent overfitting. Letting $B$ denote the set of samples in a batch during training, the overall batch loss functions with hyperparameter $\lambda$ are:

%\begin{centermath}
\begin{align}
    & L^B_{base} = \frac{1}{|B|} \sum\limits_{i\in B} L^{(i)}_Y + \frac{\lambda}{2}\norm{W}^2_F \\
    & L^B_{localize} = \frac{1}{|B|} \sum\limits_{i\in B} \left[L^{(i)}_Y + L^{(i)}_O\right] + \frac{\lambda}{2}\norm{W}^2_F
\end{align}
%\end{centermath}

\noindent The loss functions $L^B_{base}$ and $L^B_{localize}$ correspond to network configurations MPN-base and MPN-localize, respectively. MPN-base does not make use of stepwise labels and hence does not incorporate the $L_O$ loss. It is suitable for situations where stepwise labels are unavailable or label localization is not required.

\begin{figure*}
    \centering
    \begin{subfigure}{0.3\linewidth}
        \centering
        \includegraphics[width=0.95\textwidth]{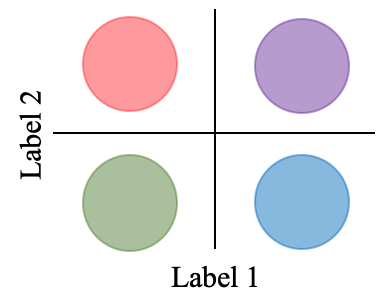}
        \caption{Target distribution}
         \label{fig:embedding}
    \end{subfigure}%
    \begin{subfigure}{0.3\linewidth}
        \centering
        \includegraphics[width=0.95\textwidth]{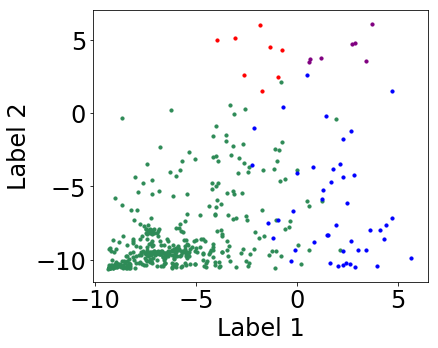}
        \caption{Learned distribution}
        \label{fig:empericalembedding}
    \end{subfigure}
    \begin{subfigure}{0.3\linewidth}
        \centering
        \includegraphics[width=0.95\textwidth]{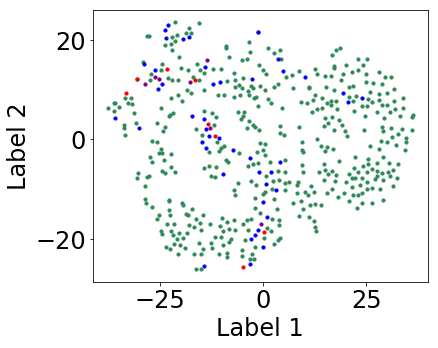}
        \caption{Original distribution}
        \label{fig:tsne}
    \end{subfigure}  
    \begin{subfigure}{0.09\linewidth}
        \includegraphics[width=0.9\textwidth]{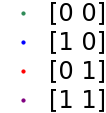}
    \end{subfigure}      
    \caption{Embedded representations $g$ and original distribution for two labels. $[n_1 \hspace{2pt} n_2]$ denotes the segment label to predict.}
\end{figure*}
\subsection{Multi-label Decision}
Figure \ref{fig:tsne} shows the t-SNE representation of the original samples of the PHM dataset for two fault types, which demonstrates that the samples are not readily separable at their original state.
Figure \ref{fig:embedding} shows the ideal distribution of $g^{(i)}$ across $i$ learned for two labels. In this representation, samples with identical label vectors are distinctly clustered together, and a threshold at zero would clearly separate samples that disagree on a given label $\ell$. Figure \ref{fig:empericalembedding} plots the actual embeddings produced by MPN, which are evidently more linearly separable than the original samples.

Cluster radius is typically governed by the data. Two observations can be made. First, the more data points in a cluster, the larger the cluster's radius. Second, the larger the noise component observed in the data, the larger the cluster radius. Both cases may lead to cluster overlap as shown in Figure\ref{fig:empericalembedding}. Thus in practice it may be difficult to immediately note the correct classification boundary. However, standard classification algorithms can be used to find this boundary. An example here is a linear one-vs-all SVM classifier which was found to be robust to input data during our experiment. Label localization, or prediction of labels per step in the future interval, can be achieved using a linear SVM classifier.

\section{Experiments using Real Data}
\label{sec:applications}
The experiments for this study compare the performance of MPN to six state-of-the-art multi-label classifiers in two applications: An ensemble of classifiers on hand-engineered features and observations (Feature); a KNN classifier using majority vote and the DTW distance (DTW); Shapelet Forests (S-F) \cite{patri2014} which require learning on a powerset of labels; a  multi-label LSTM network with replicated targets (ML-LSTM) \cite{lipton2015}; a Siamese LSTM network learned metric \cite{mcclure2017} used in a KNN classifier (Siamese-LSTM); and a SVM using the RBF kernel which does not explicitly model the temporal input (SVMrbf). Details of these methods can be found in Section \ref{sec:related}. As noted before, the inputs to these classifiers are $z'_{1:\tau}$ and $c_{1:T}$, where $z'_{1:\tau}$ is $z_{1:\tau}$ padded to the length of $c$ with the feature mean. This is a common approach taken to remedy the case of missing data from real life data collection. Because these classifiers do not learn the stepwise labels $\tilde{o}$ they are compared with MPN-base which does not have the loss component $L_O$. 

In addition, different features and ensemble of classifiers are used for each of the two applications due to differences of the two domains. Details of the datasets are in Sections \ref{sec:PHM} and \ref{sec:HAR}. The two methods are named Feature (PHM) and Feature (HAR) according to the two domains. Feature (PHM) includes all features used in the winning entry \cite{xiao2016} of the 2015 PHM Society Data Challenge, namely month, weekday, hour, day, time, and historical observations. The ensemble comprises gradient boosting machine, random forest and penalized logistic regression, which are the best-performing methods for the tasks in Xiao's paper \cite{xiao2016}. The ensemble in Feature (HAR) comprises three best-performing methods in the 2011 Opportunity Activity Recognition Challenge, namely SVM, 1-Nearest Neighbor and decision tree \cite{opportunity2011}. The corresponding features (mean, variance, maximum, minimum and historical observations) are included for Feature (HAR). Feature engineering is a common preprocessing technique for human activity recognition \cite{figo2010}. More recent methods utilize deep learning and are moving away from being feature-based \cite{ordonez2016}, but these methods are mainly focused on segmentation and classification and not prediction. 

Detailed experiments for this paper also studied the effects of different multi-label decision classifiers and auxiliary losses in MPN. The performance of three classifiers is tested. The default classifier on the embedded representations is the linear SVM. Two other classifiers tested are hard thresholding at zero and assuming the nearest class as measured by distance from the empirical class mean. MPN-base and MPN-localize are network configurations with loss functions $L^B_{base}$ and $L^B_{localize}$, respectively. Siamese MPN-localize is a Siamese version of MPN-localize with an auxiliary Siamese objective measuring the similarity $\tilde{s}^{(i,j)}[\ell]$ of label $\ell$ for sample pair $(i,j)$ through $s^{(i,j)}[\ell] = \exp{\left(-\left| g^{(i)}[\ell] - g^{(j)}[\ell] \right|\right)}$. This version is included since the Siamese structure has shown success in classification tasks with class imbalance \cite{pei2016, mueller2016, neculoiu2016}. Siamese MPN-localize is trained with loss function:
%\begin{centermath}
\begin{align}
    & L^B_{siamese} = \frac{1}{|pair(B)|} \sum\limits_{(i,j)\in pair(B)} \nonumber\\
    & \hspace{2.5cm} \left[\beta\paren{L^{(i)}_Y + L^{(j)}_Y + L^{(i)}_O + L^{(j)}_O} + \right. \nonumber\\ 
    & \hspace{2.5cm} \left. (1-\beta)L^{(i,j)}_S\right] + \frac{\lambda}{2}\norm{W}^2_F
\end{align}
%\end{centermath}

\noindent where $pair(B)$ is a set of unique pairs of samples in $B$, and $L^{(i,j)}_S$ is the squared loss for $s^{(i,j)}$:
% \begin{centermath}
% \begin{array}{ll}
%     L^{(i,j)}_S = \norm{ \tilde{s}^{(i,j)} - s^{(i,j)} }^2_2
% \end{array}
% \end{centermath}
%\begin{centermath}
\begin{align}
    & L^{(i,j)}_S = \frac{1}{L}\left[\tilde{s}^{(i,j)} \cdot \left(1-s^{(i,j)}\right)^2 + \right. \nonumber\\
    & \hspace{3cm} \left. \left(1-\tilde{s}^{(i,j)}\right) \cdot \left(s^{(i,j)}\right)^2\right]
\end{align}
%\end{centermath}
For the experiments, we train all three network configurations using $L^B_{siamese}$ with the appropriate loss components turned off, so that all configurations have the same number of trainable parameters and are comparable.

Performance is evaluated using the F1-score, precision and recall. Micro and macro averages are presented because metric choice depends on the importance of the rare labels to the user. Macro metrics average the performance of each label type, and micro metrics average the performance across each sample. 
 
For all methods, a hyperparameter grid search is performed on a held out validation set to maximize the sum of micro and macro F1-scores. The hyperparameters tuned and the search values are presented in Table \ref{table:hyperparams}. In all experiments, there are 500, 100, and 400 training, validation, and test samples, respectively. 

\begin{table}[h]
\begin{subtable}[h]{0.45\textwidth}
\centering
\begin{tabular}{ |p{2.2cm}|p{5cm}| }
 \hline
 Method & Hyperparameter \\ 
 \hline
 Feature (PHM) & Regularization parameter ($C$)\\
 Feature (HAR) & RBF kernel parameter ($\gamma$), regularization parameter ($C$)\\
 DTW & number of nearest neighbors ($k$), warping window ($w$)\\
 SVMrbf & RBF kernel parameter ($\gamma$), regularization parameter ($C$)\\
 S-F & None\\
 ML-LSTM & learning rate ($\eta$), intermediary target parameter ($\alpha$), dropout probability ($p$)\\
 Siamese-LSTM & number of nearest neighbors ($k$), learning rate ($\eta$), margin parameter ($m$), dropout probability ($p$)\\
 MPN-base & learning rate ($\eta$), regularization parameter ($\lambda$)\\
 MPN-localize & learning rate ($\eta$), regularization parameter ($\lambda$)\\
 Siamese MPN-localize & learning rate ($\ell$), weighing parameter ($\beta$), regularization parameter ($\lambda$)\\
 \hline
\end{tabular}
\caption{Hyperparameters tuned}
\end{subtable}

\hfill

\begin{subtable}[h]{0.45\textwidth}
\centering
\begin{tabular}{ |p{2.2cm}|p{5cm}| }
 \hline
 Hyperparameter & Values \\ 
 \hline
 $k$ & $1, 3, 5, 7, 9$\\
 $w$ & $0, 1, 2$\\
 $\gamma$, $C$ & $0.001, 0.01, 0.1, 1, 10$\\
 $\eta$ & $0.001, 0.01, 0.1$\\
 $\alpha$, $\beta$, $p$, $m$ & $0.1, 0.3, 0.5, 0.7, 0.9$\\
 $\lambda$ & $0.01, 0.1, 1$\\
 \hline
\end{tabular}
\caption{Values tuned}
\end{subtable}

\caption{Hyperparameters tuned for each method}
\label{table:hyperparams} 
\end{table}

\subsection{Fault Prediction for Industrial Plants}
\label{sec:PHM}

The PHM dataset contains plant data from 33 plants with labeled data for 6 types of failure events \cite{phm15}. The sampling interval  for all plants is 15 minutes. The dataset contains time series samples for several  years. For this reason fixed-length samples (for each plant) are chosen using random sampling. Sample length for historical observations is 30 (7.5 hours), and the forecast length for future events is 10 (2.5 hours). These durations are chosen because the dataset description stipulates that faults are predictable at most three hours in advance. Observation input $z$ contains sensor measurements and contextual input $c$ contains control system setpoints, and their combined dimension is 20 to 116 across plants. Because each plant has a unique sensor set, testing and training is performed separately for each plant. 

\begin{figure}
    \centering
    \includegraphics[width=\linewidth]{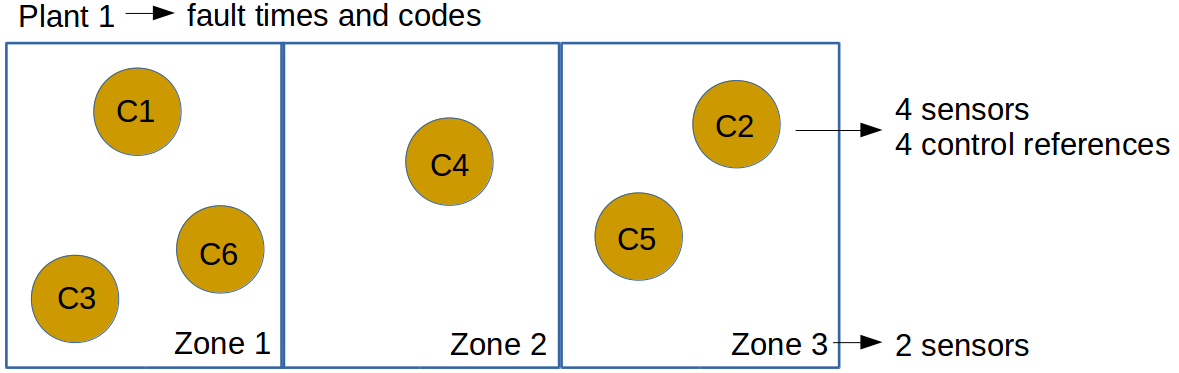}
    \caption{PHM: Monitoring setup for Plant 1 with components (yellow circles) split over zones. True component-to-zone mapping is not provided, and included only for illustration purposes.}
    \label{fig:plant1setup}
\end{figure}

Each plant has a particular number of monitored components, and each component corresponds to four sensors and four control references. The components are further grouped by zones, and there are two sensors providing environmental readings per zone. The true component-to-zone mapping is not provided in the dataset, hence no mapping is explicitly enforced through model input. Fault instances are given in terms of a start and end time, and a code specifying the fault type. This information is formatted according to our multi-label objective as $\tilde{y}$ and $\tilde{o}_{\tau+1:T}$ for training and testing the models. An example experiment setup for Plant 1 is shown in Figure \ref{fig:plant1setup}. Plant 1 has six components split over three zones.

\begin{figure*}
    \centering
    \begin{subfigure}{0.8\linewidth}
        \centering
        \includegraphics[width=\textwidth]{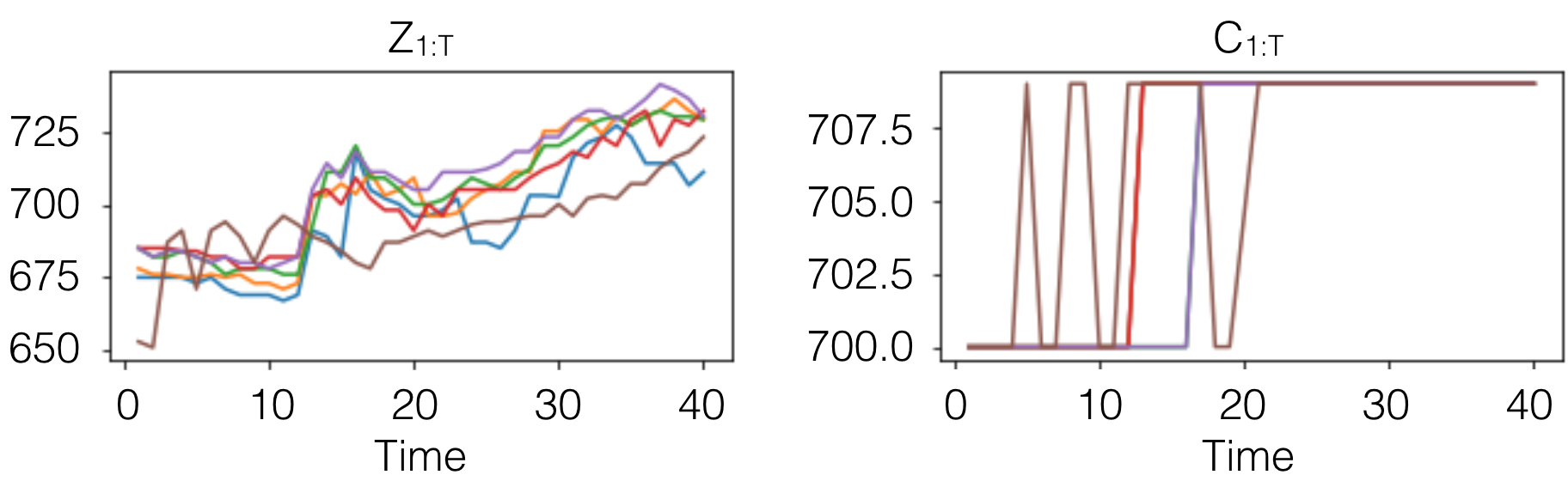}
        \caption{\label{fig:PHMsamp1} Sample with no faults}
    \end{subfigure}%
    
    \begin{subfigure}{0.8\linewidth}
        \centering
        \includegraphics[width=\textwidth]{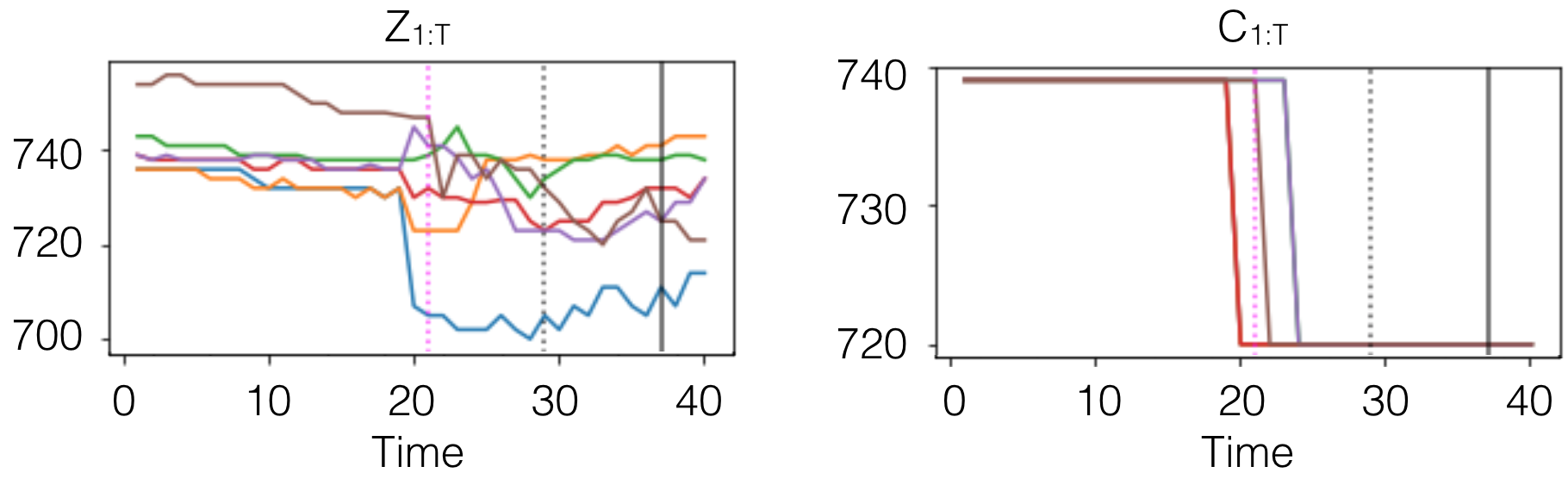}
        \caption{\label{fig:PHMsamp2} Sample with fault labels 1 and 6. Dotted and solid vertical line mark the start and end of a fault, respectively. Black denotes fault label 1 and pink denotes fault label 6.}
    \end{subfigure}
    \caption{PHM: Plant 1 sample measurements of the first sensor and first control reference across six components.}
    \label{fig:PHMsamples}
\end{figure*}

This dataset has extreme label imbalance with important events that happen infrequently. This is demonstrated in the label distribution table in Table \ref{table:plant1}, where fault label 4 is observed only 8 times in the 600 training and validation samples of Plant 1. Furthermore, multiple concurrent faults can create unique system states. Figure \ref{fig:PHMsamples} shows two samples of measurements from the first sensor and first control reference across the six components monitored in Plant 1. Figure \ref{fig:PHMsamp1} comes from a healthy sample with no faults. The sensor readings demonstrate an upward trend and fluctuations, possibly due to frequent changes of the setpoints in the control reference. This indicates the importance of including information of future context for prediction. Figure \ref{fig:PHMsamp2} comes from a sample with co-occurring fault labels 1 and 6. Fault 6 starts in the middle and persists till the end of the observed time frame, and Fault 1 starts and ends in the second half of the time frame. Multi-label objectives are required due to the variety of label combinations.

\begin{table}
\centering
\begin{tabular}{ |c|c| } 
\hline
 Fault Code & Occurrences   \\ 
 \hline\hline
 Code 1 & 58  \\ 
 Code 2  & 43  \\ 
 Code 3  & 31  \\ 
 Code 4 & 8  \\ 
 Code 5  & 20  \\ 
 Code 6  & 155  \\ 
 \hline
 
\end{tabular}
\caption{\label{table:plant1}PHM: Number of Plant 1 training samples per fault label.}
\end{table}

\subsubsection{Results and Discussion}
\label{sec:PHM results}

\begin{figure*}
    \centering
    \includegraphics[width=0.9\linewidth]{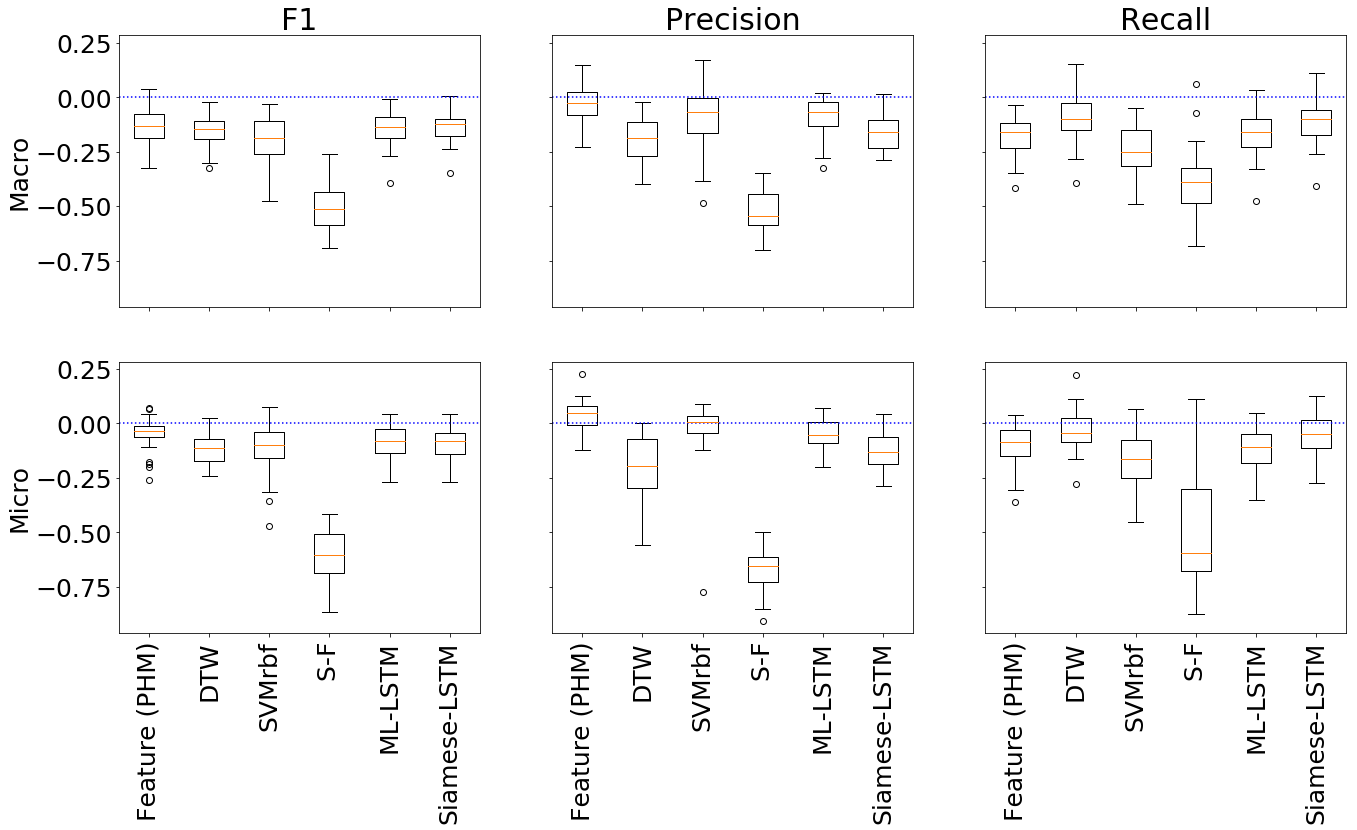}
    \caption{PHM: Label prediction test performance of methods \textit{subtracted by MPN-base} for each of 33 plants. Negative value indicates that MPN-base outperformed in the metric for the plant.}
    \label{fig:plantpred}
\end{figure*}

Figure \ref{fig:plantpred} shows the multi-label prediction test performance, where the performance of MPN-base for each plant is subtracted from that of competing methods for the corresponding plant. The illustration allows direct comparison of the methods for each of the 33 plants. A negative value after subtraction indicates a superior performance of MPN-base, which is mostly the case at both micro and macro levels for precision, recall and F1-scores. The hyperparameters most frequently selected for each model across the 33 plants are presented in Table \ref{table:hyperparam(PHM)}.
\begin{table}
\centering
\begin{tabular}{ |p{2.2cm}|p{5cm}| }
 \hline
 Method & Selected hyperparameter\\ 
 \hline
 Feature (PHM) & $C = 0.1$\\
 DTW & $k = 1$, $w = 2$\\
 SVMrbf & $\gamma = 0.001$, $C = 10$\\
 S-F & None\\
 ML-LSTM & $\eta = 0.001$, $\alpha = 0.1$, $p = 0.9$\\
 Siamese-LSTM & $k = 1$, $\eta = 0.001$, $m = 0.1$, $p = 0.9$\\
 MPN-base & $\eta = 0.001$, $\lambda = 1$\\
 MPN-localize & $\eta = 0.01$, $\lambda = 0.1$\\
 Siamese MPN-localize & $\eta = 0.01$, $\beta = 0.3$, $\lambda = 0.1$\\
 \hline
\end{tabular}
\caption{PHM: Hyperparameters most frequently selected across 33 plants.}
\label{table:hyperparam(PHM)}
\end{table}

S-F performed the worst, because the limited number of samples corresponding to each label combination does not allow sufficiently diverse shapelets to be extracted for classification. The other five methods performed similarly with SVMrbf being slightly worse overall since it does not attempt to capture the dynamics of time series. Feature (PHM) has slightly higher micro-F1 scores but lower macro-F1 scores, which implies that it has worse prediction performance for rare labels.
In general, the competing methods did not perform as well as MPN-base, possibly because they are not constructed to make effective use of future context. The additional future context results in an irregular input data shape. Padding of the samples, a common remedy for missing data, allows the inclusion of the extra information in these models, but padding with a constant can cause a bias in learning since the missing data is not random and the data is temporal in nature.

\begin{figure}
    \centering
    \includegraphics[width=0.4\textwidth]{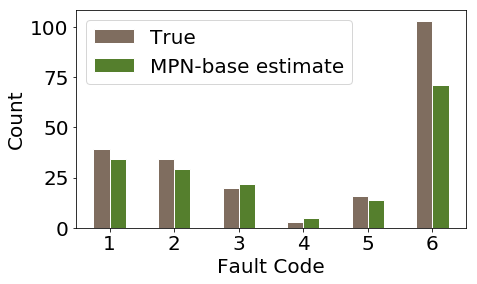}
    \caption{\label{fig:plant1test}PHM: Number of true and estimated Plant 1 test samples per fault label.}
\end{figure}

Although some training labels are rare as the Plant 1 example shows in Table \ref{table:plant1}, the proposed MPN-base is able to learn the full variety of labels. Figure \ref{fig:plant1test} shows the distribution of the true Plant 1 test labels versus the labels predicted by MPN-base. The counts are similar across all labels, and MPN-base is capable of predicting label 4 despite it being underrepresented in the training set.

{\bf Effects of model design:} Table \ref{table:faultclassifier} shows three different methods of classification on the embedded representations: linear SVM; hard thresholding at zero; and assuming the nearest class as measured by distance from the empirical class mean. The SVM shows the best overall performance. Note the poor performance of the third approach is because the variance of the representations is different across the classes.
 
\begin{table}
\centering
\begin{tabular}{|l|l|l|l|l|}
\hline
\multirow{2}{*}{Classifier} & \multicolumn{2}{l|}{Macro-F1} & \multicolumn{2}{l|}{Micro-F1} \\ \cline{2-5} 
                            & Mean         & Var       & Mean         & Var       \\ \hline
Linear SVM                  & 0.610        & 0.017          & 0.774        & 0.007          \\ \hline
Threshold at 0              & 0.603        & 0.018          & 0.765        & 0.007          \\ \hline
Nearest class               & 0.599        & 0.019          & 0.590        & 0.044          \\ \hline
\multirow{2}{*}{Network}    & \multicolumn{2}{l|}{Macro-F1} & \multicolumn{2}{l|}{Micro-F1} \\ \cline{2-5} 
                            & Mean         & Var       & Mean         & Var       \\ \hline
MPN-base                    & 0.610        & 0.017          & 0.774        & 0.007          \\ \hline
MPN-localize                & 0.605        & 0.021          & 0.771        & 0.007          \\ \hline
Siamese MPN-localize        & 0.603        & 0.020          & 0.773        & 0.008          \\ \hline
\end{tabular}
% \begin{tabular}{|l|l|l|l|l|}
% \hline
% \multirow{2}{*}{Classifier} & \multicolumn{2}{l|}{Macro-F1} & \multicolumn{2}{l|}{Micro-F1} \\ \cline{2-5} 
%                             & Mean         & Var       & Mean         & Var       \\ \hline
% Linear SVM                  & 0.513        & 0.018          & 0.715        & 0.015          \\ \hline
% Threshold at 0              & 0.509        & 0.018          & 0.712        & 0.012          \\ \hline
% Nearest class               & 0.504        & 0.019          & 0.499        & 0.052          \\ \hline
% \multirow{2}{*}{Network}    & \multicolumn{2}{l|}{Macro-F1} & \multicolumn{2}{l|}{Micro-F1} \\ \cline{2-5} 
%                             & Mean         & Var       & Mean         & Var       \\ \hline
% MPN-base                    & 0.513        & 0.018          & 0.715        & 0.015          \\ \hline
% MPN-localize                & 0.517        & 0.019          & 0.720        & 0.014          \\ \hline
% Siamese MPN-localize        & 0.509        & 0.020          & 0.713        & 0.015          \\ \hline
% \end{tabular}
\caption{PHM: Performance of MPN-base network given choice of classifier, and performance of three versions of the MPN network. The mean values are averaged across 33 plants.}
\label{table:faultclassifier} 
\end{table}

Table \ref{table:faultclassifier} shows a comparison of MPN-base, MPN-localize and Siamese MPN-localize. All three versions have similar F1-scores, showing that the auxiliary Siamese targets added in this experiment do not improve learning. Separation of the embedded representations across different labels is achieved sufficiently by the loss functions $L^B_{base}$ and $L^B_{localize}$.

\begin{figure}
    \centering
    \includegraphics[width=0.8\linewidth]{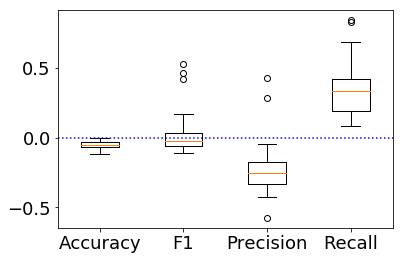}
    \caption{PHM: Fault localization test performance of multi-label prediction baseline subtracted by label localization for each of 33 plants.}
    \label{fig:plantlocalization}
\end{figure}

{\bf Fault localization:} For testing the label localization performance of MPN-localize, the baseline is the multi-label prediction output without temporal localization, i.e. a fault label is either present/absent at all forecast time steps. Figure \ref{fig:plantlocalization} shows the localization performance subtracted from the baseline performance for each of the 33 plants. As expected, the baseline has higher recall but also more false positives. Overall, localization improves accuracy and the F1-score.

\subsection{Label Prediction for Human Activity Recognition}
\label{sec:HAR}

This application of human activity prediction is included to further demonstrate the effectiveness of the proposed MPN in multi-label prediction, and also its usage outside of PHM domains.

The Opportunity Dataset contains time series data used for HAR tasks \cite{roggen1020}. Each of 4 subjects perform an activity of daily living (ADL) 5 times.
Sensor readings are collected from sensing systems deployed on the body of the subjects, on objects and in the environment.
The ADL consists of a sequence of high-level activities, namely early morning, coffee time, sandwich time, cleanup and relaxing. On a finer level, each action is described by the motion of the arms and the objects the arms interact with. We are interested in predicting the fine-level motions based on the sensor readings and high-level activities.

In this experiment, fixed-length samples are uniformly sampled with observed and forecast lengths as 75 and 25, respectively. The observation input $z$ is the 243-dimensional measurements from GPS, motion sensors and other sensing units, and the contextual input $c$ is the indicator matrix for high-level activities which empirically can be inferred from the time of day and the subject's pattern of past schedules.
Labels to be predicted are the low-level activities of the right arm. This is a combination of its motion in 13 categories including open and unlock, and the object the arm is interacting with which consists 23 options such as bottle and dishwasher. That is, $L = 36$, Label occurrences range from 1 to 180 times. This is a challenging task due to label sparsity and the inherent difficulty of predicting human activities \cite{Ryoo2011}. 

\subsubsection{Results and Discussion}

\begin{figure*}
    \centering
    \includegraphics[width=0.9\linewidth]{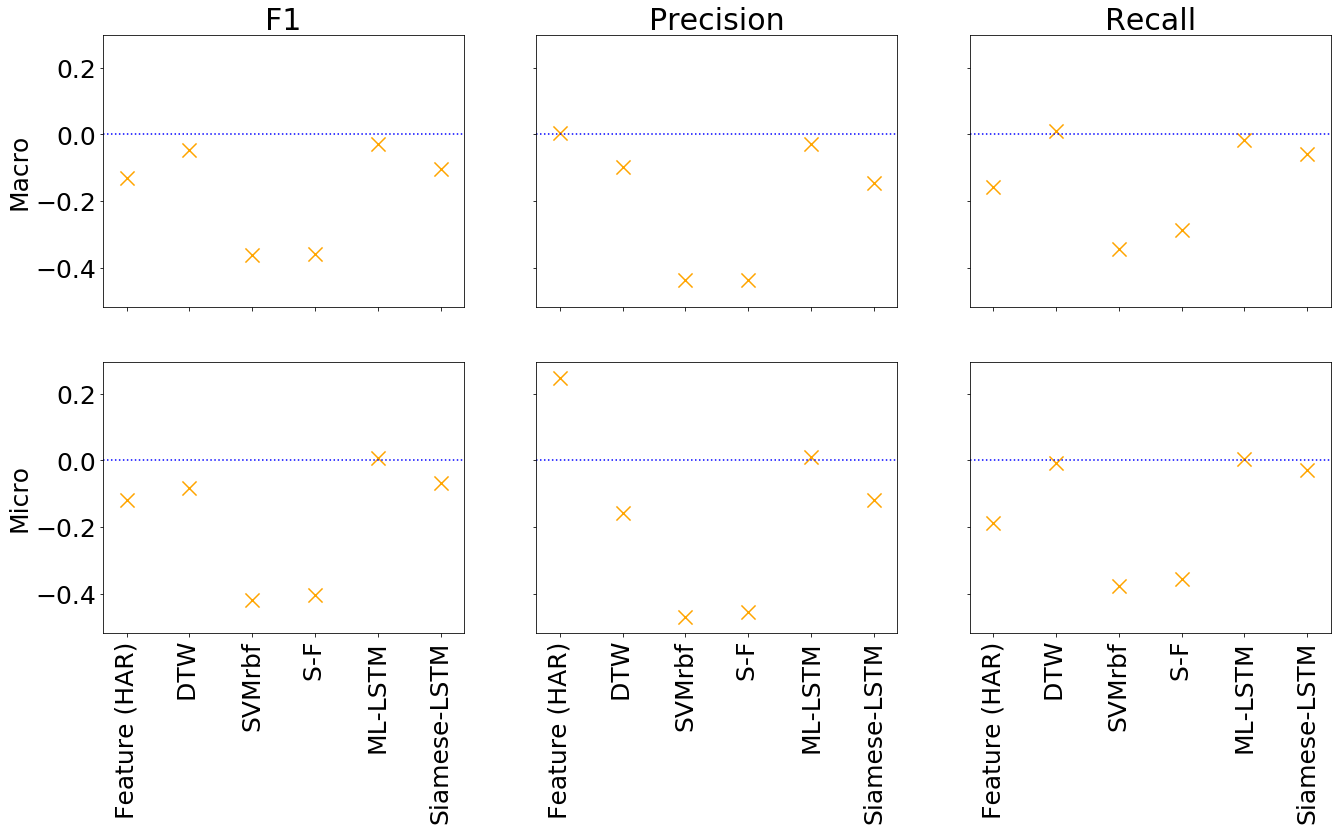}
    \caption{HAR: Label prediction test performance of methods \textit{subtracted by MPN-base}. Negative value indicates that MPN-base outperformed in the metric.}
    \label{fig:harpred}
\end{figure*}

Figure \ref{fig:harpred} shows the test performance of each competing method subtracted by that of MPN-base. The proposed method has the best overall performance. This is similar to the results in the PHM application in Section \ref{sec:PHM results}.

The hyperparameters selected for each model are presented in Table \ref{table:hyperparam(HAR)}.
\begin{table}
\centering
\begin{tabular}{ |p{2.2cm}|p{5cm}| }
 \hline
 Method & Selected hyperparameter \\ 
 \hline
 Feature (HAR) & $\gamma = 0.001$, $C = 0.001$\\
 DTW & $k = 1$, $w = 2$\\
 SVMrbf & $\gamma = 10$, $C = 0.1$\\
 S-F & None\\
 ML-LSTM & $\eta = 0.001$, $\alpha = 0.1$, $p = 0.9$\\
 Siamese-LSTM & $k = 1$, $\eta = 0.01$, $m = 0.1$, $p = 0.7$\\
 MPN-base & $\eta = 0.001$, $\lambda = 1$\\
 MPN-localize & $\eta = 0.01$, $\lambda = 1$\\
 Siamese MPN-localize & $\ell = 0.01$, $\beta = 0.3$, $\lambda = 0.1$\\
 \hline
\end{tabular}
\caption{HAR: Hyperparameters selected.}
\label{table:hyperparam(HAR)}
\end{table}

{\bf Effects of model design:} Table \ref{table:harclassifier} shows similar performance for SVM and thresholding, indicating that the learned representations can be classified by the zero cutoff as designed. MPN-localize has the best F1-score, showing that MPN benefits from the additional stepwise label targets in this application.

\begin{table}
\centering

\begin{tabular}{ |l|l l| }
 \hline
 Classifier & Macro-F1 & Micro-F1 \\ 
 \hline
 Linear SVM & 0.361 & 0.417 \\ 
 Threshold at 0 & 0.362 & 0.413 \\ 
 Nearest class & 0.343 & 0.298 \\
 \hline
 Network & Macro-F1 & Micro-F1 \\ 
 \hline
 MPN-base & 0.361 & 0.417 \\ 
 MPN-localize & 0.411 & 0.426 \\ 
 Siamese MPN-localize & 0.344 & 0.420 \\
 \hline
\end{tabular}
\caption{HAR: Performance of MPN-base network given choice of classifier, and performance of three versions of the MPN network.}
\label{table:harclassifier} 
\end{table}

\section{Conclusion}
\label{sec:conclusion}

This paper proposes a recurrent neural network-based approach for multi-label prediction problem in high-dimensional time series data with severe class imbalance. The proposed network, which we refer to as multi-label predictive network (MPN), embeds time series samples into a target space where the data is linearly separable. The loss function of MPN allows a variable number of labels for each sample and appropriately weighs each class to compensate for the rarity of some labels. Co-occurring labels and rare labels are common characteristics in fault prediction applications. The proposed algorithm for fault prediction using MPN performed well against state-of-the-art techniques on two different benchmark datasets, and is applicable to prediction tasks beyond PHM domains as well. 

% references section

\bibliographystyle{apacite}
\bibliography{refs}

\end{document}